\documentclass[acmsmall,screen]{acmart}
\AtBeginDocument{%
  }

\setcopyright{acmlicensed}
\copyrightyear{2026}
\acmYear{2026}
\acmDOI{XXXXXXX.XXXXXXX}

\acmJournal{TKDD}
\acmVolume{20}
\acmNumber{0}
\acmArticle{000}
\acmMonth{00}

\begin{document}

%%
%% The "title" command has an optional parameter,
%% allowing the author to define a "short title" to be used in page headers.
\title{Geometric Morphology Learning: Empowering Foundation Models from Structure to Spacetime Intelligence}

%%
%% The "author" command and its associated commands are used to define
%% the authors and their affiliations.
%% Of note is the shared affiliation of the first two authors, and the
%% "authornote" and "authornotemark" commands
%% used to denote shared contribution to the research.

\author{Li Sun}
\affiliation{%
  \institution{Beijing University of Posts and Telecommunications}
  \city{Beijing}
  \country{China}
  }
\email{lsun@bupt.edu.cn}

\author{Philip S. Yu}\correspondingauthor
\affiliation{%
  \institution{University of Illinois Chicago}
  \city{Chicago}
  \country{USA}
  }
\email{psyu@uic.edu}

%%
%% By default, the full list of authors will be used in the page
%% headers. Often, this list is too long, and will overlap
%% other information printed in the page headers. This command allows
%% the author to define a more concise list
%% of authors' names for this purpose.
\renewcommand{\shortauthors}{Li Sun and Philip Yu}

%%
%% The abstract is a short summary of the work to be presented in the
%% article.
\begin{abstract}
% The geometry is the very form that connects spaces, time and matters. 
% To bridge this chasm, 
%\vspace{-0.04in}
Geometry is the very form that connects space, time, and matter, serving as the fundamental fabric of the physical world, from continuous fields to discrete structures. Despite its importance, deep learning is inherently blind to geometry. Riemannian geometry stands as a well-established mathematical foundation for studying geometry, yet it remains largely incompatible with deep learning architectures. To bridge this gap, we establish a new learning framework---Geometric Morphology Learning---that learns geometric or topological structures, shapes or conformations for representation learning, reasoning and action modeling. Moving beyond early attempts limited to specific geometric priors, this theory centers on modeling the intrinsic geometry of objects and learning from such geometric or topological knowledge. Specifically, we construct a neural bundle, a unified scaffolding linking local and global structures, and formulate differentiable operators, including connection and curvature, thereby enabling the study of geometry with advanced geometric concepts such as holonomy and cohomology. Consequently, Geometric Morphology Learning opens new horizons in foundation models, reasoning and planning, graph modeling, physical simulation, PDE systems, and dynamics, marking a fundamental shift toward instilling systematic geometry in artificial intelligence.
\end{abstract}

%%
%% The code below is generated by the tool at http://dl.acm.org/ccs.cfm.
%% Please copy and paste the code instead of the example below.
%%
\begin{CCSXML}
<ccs2012>
<concept>
<concept_id>10010147.10010257.10010258</concept_id>
<concept_desc>Computing methodologies~Learning paradigms</concept_desc>
<concept_significance>500</concept_significance>
</concept>
<concept>
<concept_id>10010147.10010178.10010187</concept_id>
<concept_desc>Computing methodologies~Knowledge representation and reasoning</concept_desc>
<concept_significance>300</concept_significance>
</concept>
<concept>
<concept_id>10002951.10003227.10003351</concept_id>
<concept_desc>Information systems~Data mining</concept_desc>
<concept_significance>500</concept_significance>
</concept>
</ccs2012>
\end{CCSXML}

\ccsdesc[500]{Computing methodologies~Learning paradigms}
\ccsdesc[300]{Computing methodologies~Knowledge representation and reasoning}
\ccsdesc[500]{Information systems~Data mining}

% \received{20 February 2007}
% \received[revised]{12 March 2009}
% \received[accepted]{5 June 2009}

%%
%% This command processes the author and affiliation and title
%% information and builds the first part of the formatted document.
\maketitle

\begin{quote}
\centering
\textit{Manifolds are the very language of the cosmos, \\
and geometry, the ultimate cipher of spacetime.}
\end{quote}

\section{Why Geometry Matters}
% Geometry and topology form the cognitive cornerstones of both the physical and logical worlds, where objects are endowed with meaning through their spatiotemporal relationships. In chemistry and material science, the geometric arrangement of atoms directly determines the functionality, for instance, molecular chirality is critical in drug design, as different enantiomers can yield drastically different therapeutic effects, while lattice topology such as Kagome lattice determines the electronic character of the material. In biology, functional mechanisms of proteins such as folding, cavity formation, and binding are deeply intertwined with intrinsic topological structures. In physical, the behavior of fluid dynamics is strictly governed by the geometric boundaries of the environment, such as the flow of air over an aircraft wing or water through a complex vascular network. Solving their Partial Differential Equations (PDEs) without obeying the geometric constraints leads to catastrophic consequences and invalid results. Another profound example of geometry’s primacy is found in physical cosmology, the gravity is not merely a force but a manifestation of the curvature of spacetime itself and this geometry, in turn, dictates the motion of matter. That is, the geometric structure is the very fabric of the reality.

%molecular chirality—the geometric property of a molecule that is non-superimposable on its mirror image

Geometry and topology form the cognitive cornerstones of both the physical and logical realms, endowing objects with meaning through their spatiotemporal relationships. In chemistry and materials science, the geometric arrangement of atoms directly dictates functionality, for example, 
molecular chirality\footnote{Molecular chirality refers to the geometric property of a molecule that is non-superimposable on its mirror image.} is critical in drug design, 
%molecular chirality is critical in drug design, 
as different enantiomers can yield drastically different therapeutic effects \cite{Chirality}. Similarly, lattice topology, such as that of the Kagome lattice, fundamentally determines a material’s electronic properties \cite{material}. In biology, the functional mechanisms of proteins, including folding, cavity formation and molecular binding, are inextricably intertwined with the intrinsic topological structures \cite{aFold3}. In physics, fluid dynamics is strictly governed by the geometric boundaries of the environment, whether involving airflow over an aircraft wing or water flowing through a complex vascular network. Solving the associated partial differential equations (PDEs) without adhering to these geometric constraints leads to catastrophic failures and invalid results. 

Understanding geometric structures remains a formidable challenge for modern artificial intelligence. Conventional deep learning implicitly assumes that data resides in a flat, Euclidean space, where vector operations are globally valid. Consequently, it conflates extrinsic distances (straight lines in the representation space) with intrinsic geodesics (the shortest paths along a data manifold). This fundamental mismatch causes curved manifolds, such as 3D meshes or proteins, to undergo distortion or twisting when mapped onto flat-space representations. Furthermore, such coordinate-dependent representation learning fails to capture intrinsic, coordinate-invariant properties. Although geometric deep learning addresses this limitation by  emphasizing equivariance \cite{GDL}, it remains largely blind to the intrinsic geometric fabric of the data. While Riemannian geometry offers a rigorous mathematical framework, the absence of differentiable formulations for its core concepts impedes their direct integration into data-driven neural networks.

%\vspace{-0.05in}
\subsubsection*{\textbf{Next-Generation Intelligence.}}
% We propose Geometric Morphology Learning, a new learning framework that aligns machine intelligence with the very fabric of reality---intrinsic geometric or topological structures. 
% It encourages a learning paradigm shift from the reliance on specific geometric priors or focus on certain pattern of interest to model the geometric morphology locally and globally for knowledge discovery. 
% We establish the framework, theory and formulation for geometric morphology learning so that neural networks perceive, reason, and act  with the principles of modern mathematics.

We propose Geometric Morphology Learning, a new learning framework that aligns machine intelligence with the intrinsic geometric and topological structures of reality. This framework catalyzes a fundamental paradigm shift, moving beyond the use of specific geometric priors  towards modeling geometric morphology across both local and global scales for knowledge discovery. To achieve this, we establish the theoretical foundations and mathematical formulations of this framework, empowering neural networks to perceive, reason, and act in accordance with the principles of modern mathematics.
% \begin{itemize}
%    \item \textbf{A Paradigm Shift.}
%    \item \textbf{Theory \& Formulations.}
% \end{itemize}

%%\vspace{-0.05in}
\section{Riemannian Geometry: Opportunities and Challenges}

%\vspace{-0.02in}
%Conventional neural networks are inherently blind to the geometry of data, whereas Riemannian geometry offers the opportunity to instill systematic geometric awareness into artificial intelligence.

Riemannian geometry offers a solid mathematical framework but is incompatible with deep learning.

%\vspace{-0.05in}
%\subsubsection*{\textbf{An Elegant Mathematical Discipline.}}
\subsubsection*{\textbf{A Mathematical Discipline.}}
% Riemannian geometry bridges the annals of science and the modern era. 
Riemannian geometry bridges the legacy of classical science and the frontiers of the modern era. 
Its history was ignited by a simple yet revolutionary question in 1850s: Through a point not on a given line, how many lines can be drawn parallel to the given line? 
% Its brilliant history starts by a simple question in 1850s: Through a point not on a given line, how many lines can be drawn parallel to the given line?
By challenging Euclid’s Parallel Postulate, Bernhard Riemann unveiled a breathtaking new landscape of geometry.
Building upon the invaluable legacy of pioneering mathematicians\footnote{
These great mathematicians contributed the fundamentals in Riemannian geometry, including Elwin B. Christoffel’s symbols and tensor calculus, Élie Cartan’s exterior derivative and moving frames, Hermann Weyl and Tullio Levi-Civita’s deep explorations into connections, and Shiing-Shen Chern’s fiber bundles and characteristic classes, etc.},
 Riemannian geometry has evolved into an elegant and rigorous framework for studying spaces, structures, shapes, and the very fabric of the universe.
% Nowadays, it remains the indispensable core of modern mathematics, acting as the vital engine behind Shing-Tung Yau’s Geometric Analysis and Shigefumi Mori’s work in Algebraic Geometry. Beyond pure mathematics, it serves as the fundamental language of modern physics, e.g., in Albert Einstein’s General Relativity, the gravity is understood as the curvature of spacetime. It also provides the essential geometric framework for Quantum Mechanics and Superstring Theory, where the universe’s hidden dimensions are modeled as complex Calabi-Yau manifolds. 
Today, it remains an indispensable cornerstone of modern mathematics as exemplified by Shing-Tung Yau’s  work \cite{yau1974curvature} in geometric analysis and Shigefumi Mori’s work \cite{mori1988flip} in algebraic geometry. Beyond pure mathematics, it serves as a fundamental language of modern physics most notably in Albert Einstein’s  general theory of relativity and in gauge theory \cite{geo-book}. 
In essence, Riemannian geometry continues to shape our deepest understanding of the abstract mathematical universe and physical reality.

%\vspace{-0.05in}
\subsubsection*{\textbf{Fundamental Concepts.}}

The fundamental object is the manifold \cite{manifold-book}, a topological space that locally resembles Euclidean space but may exhibit globally complex, nonlinear structures. 
This local resemblance is formalized through the concept of a coordinate chart, 
%which maps open subsets of the manifold to Euclidean space, 
thereby enabling calculus on curved domains. 
%When equipped with a Riemannian metric tensor to measure distances and angles, 
When equipped with a Riemannian metric, 
a manifold admits a unique Levi-Civita connection, which in turn gives rise to Ricci curvature tensor. 
%This tensor quantifies the deviation in the volume of a geodesic ball in a curved manifold relative to that in standard Euclidean space, essentially characterizing geometric distortion and, in the context of General Relativity, gravitational effects.
This tensor quantifies  infinitesimal deviations  in the volume of a geodesic ball in a curved manifold from its counterpart  in standard Euclidean space, thereby characterizing geometric distortion.
By constructing infinitesimal geometry via tangent spaces, Hermann Weyl laid the foundation for studying geometry using  differentials, where
an concept of continuity  abstracted as a connection.\footnote{A connection allows vectors to be compared across different points and defines the rules of parallel transport. } 
Concurrently, Lie algebras provide an algebraic framework for describing these geometric structures,
%Concurrently, Lie algebras provide a framework for infinitesimal symmetries and transformations governing these geometric structures. 
while diffeomorphisms ensure geometric invariance under smooth deformations. 
Élie Cartan introduced a  local perspective through the moving frame method and exterior derivative; this coordinate-free method elucidates the concept of holonomy.\footnote{The concept of holonomy characterizes the twisting  by tracking the evolution of a vector traversing a closed loop.}
% Complementing this local analysis, Shiing-Shen Chern integrated local geometry with fiber bundles, providing a topological scaffolding that extends localities to global topology. 
% Formulated through cohomology theories, Chern classes serve as topological invariant which measures the obstructions to defining global frames on these bundles.
Shiing-Shen Chern unified local and global analysis through fiber bundles, intertwining geometry and topology. 
%A recent development by Shing-Tung Yau is the integration between  geometric analysis with nonlinear PDEs
Subsequently, Shing-Tung Yau bridged geometric analysis and nonlinear PDEs
\footnote{A noteworthy achievement in this field is the discovery of the Calabi-Yau manifold \cite{yau1977calabi}, a solution to the Einstein equations. The groundbreaking proof of its existence earned Shing-Tung Yau the Fields Medal, one of the highest honors in mathematics.}, enriching the analytical tools available for studying these structures.

%--smooth, invertible mappings that ensure intrinsic geometric properties remain invariant under smooth deformations. 

%\vspace{-0.05in}
\subsubsection*{\textbf{Challenges for Deep Learning.}}
% The deep learning trains a neural network through gradient backpropagation, and thus the differentiablity is a pre-requirement of neural architecture.
% However, except for some special cases\footnote{The distant metric and Lie algebra in hyperbolic spaces.}, the key concepts for general geometry, such as fiber, connection and transport, are given as abstract establishment and lack analytical forms or differentiable instantiations.  

% Thus it calls for a new systematic constructions to enable deep learning with geometry.

% Riemannian geometry is generally incompatible with deep learning. Deep learning trains neural networks via gradient backpropagation, and thus the differentiability is a fundamental prerequisite for neural architectures. However, with the exception of specific cases\footnote{Distance metrics and Lie algebras in hyperbolic spaces.}, the key concepts for studying geometry, e.g., fiber bundles, connections, and parallel transport, are formulated abstractly and lack differentiable instantiations. Existing deep models applying Riemannian geometry  often restrict themselves to off-the-shelf formulations, rendering them incapable of modeling data objects with general geometric structures []. Thus, there is a pressing need for novel, systematic constructions to instill geometry into deep learning.

Riemannian geometry is generally considered incompatible with deep learning. Since deep learning relies on gradient backpropagation for model training, differentiability is a fundamental prerequisite for neural architectures. However, with the exception of specific cases\footnote{Distance metrics and Lie algebras such as exponential and logarithmic maps in hyperbolic spaces.}, key geometric concepts--such as fiber bundles, connections, and parallel transport--are formulated abstractly and lack differentiable instantiations. Existing neural architectures that apply Riemannian geometry are often restricted  to off-the-shelf formulations, rendering them incapable of modeling data with general geometric structures \cite{RGL}. Therefore, there is a pressing need to effectively instill geometric principles into deep learning.

%\vspace{-0.07in}
\section{Geometric Morphology Learning: A New Learning Framework}
%\vspace{-0.03in}
In this paper, we introduce a new learning framework---Geometric Morphology Learning, addressing geometric deficiency of deep learning while endowing the learning ability of Riemannian geometry.

% , which bridges two distinct realms of Riemannian geometry and neural networks. We address geometric deficiency of deep learning, and pioneer neural parametrizations of abstract geometry concepts, allowing for data-driven learning of geometric structures.
% This framework instills systematic geometric awareness into artificial intelligence, catalyzing a fundamental shift toward modeling intrinsic geometry and learning in accordance with modern mathematics principles.

%\vspace{-0.05in}
\subsubsection*{\textbf{A Paradigm Shift.}}
% Deep learning itself is unaware of data geometry, thus a majority solution is enforcing  deep models working within a  specific type of manifold. 
% There exist two kinds of manifolds frequently used in the context of deep learning: matrix manifold (Symmetric Positive Definite (SPD) manifold family\footnote{SPD manifolds are typical non-positive, variable curvature manifolds.}, Grassmann manifold for learning images, and Stefied manifold vital for optimization), and vector manifold which is extensively used in graph modeling in recent years.
% for example, the hyperbolic space, a negative curved, constant curvature manifold, is widely used for modeling the hierarchical data. 
% More often than not, real-world data is complex while hyperbolic space is presented as the uniform ideal structures. To alleviate this gap, recent studies introduce Cartesian product and quotient technique for representation space construction.
% They still rely on predefined geometric prior  and fitting data in such geometry, leading to the intrinsic geometric fabric important to knowledge discovery distorted. 
% Instead, we are devoted to uncovering the geometric fabric of reality. 

%it struggles to accommodate the complex, variable-curvature nature of real-world data.
Current deep learning is inherently oblivious to data geometry, prompting existing methods to embed data in predefined manifolds, such as matrix manifolds (e.g., SPD\footnote{SPD refers to the non-positive, Symmetric Positive Definite manifolds.}, Grassmann, or Stiefel manifolds) and vector manifolds (e.g., hyperbolic space). For instance, hyperbolic geometry imposes a uniform, constant-negative-curvature prior for modeling hierarchical structures \cite{ganea2018hyperbolic}, yet it frequently fails to capture the complex, variable-curvature nature of real-world data. Although recent techniques, such as Cartesian products and quotients, attempt to bridge this gap, they continue to rely on rigid geometric assumptions, thereby distorting the intrinsic geometric fabric essential for knowledge discovery. To address this fundamental flaw, our framework breaks free from predefined constraints. 
%Rather than forcing data into idealized, uniform structures, we are dedicated to directly uncovering the intrinsic geometric and topological fabric of reality, thereby performing deep learning aligned with these fundamental structures.
Rather than forcing data into idealized structures, we catalyze a fundamental shift toward directly modeling their intrinsic geometry.

\begin{figure}[t]
    \centering
        \vspace{-0.1in}
    \includegraphics[width=1\linewidth]{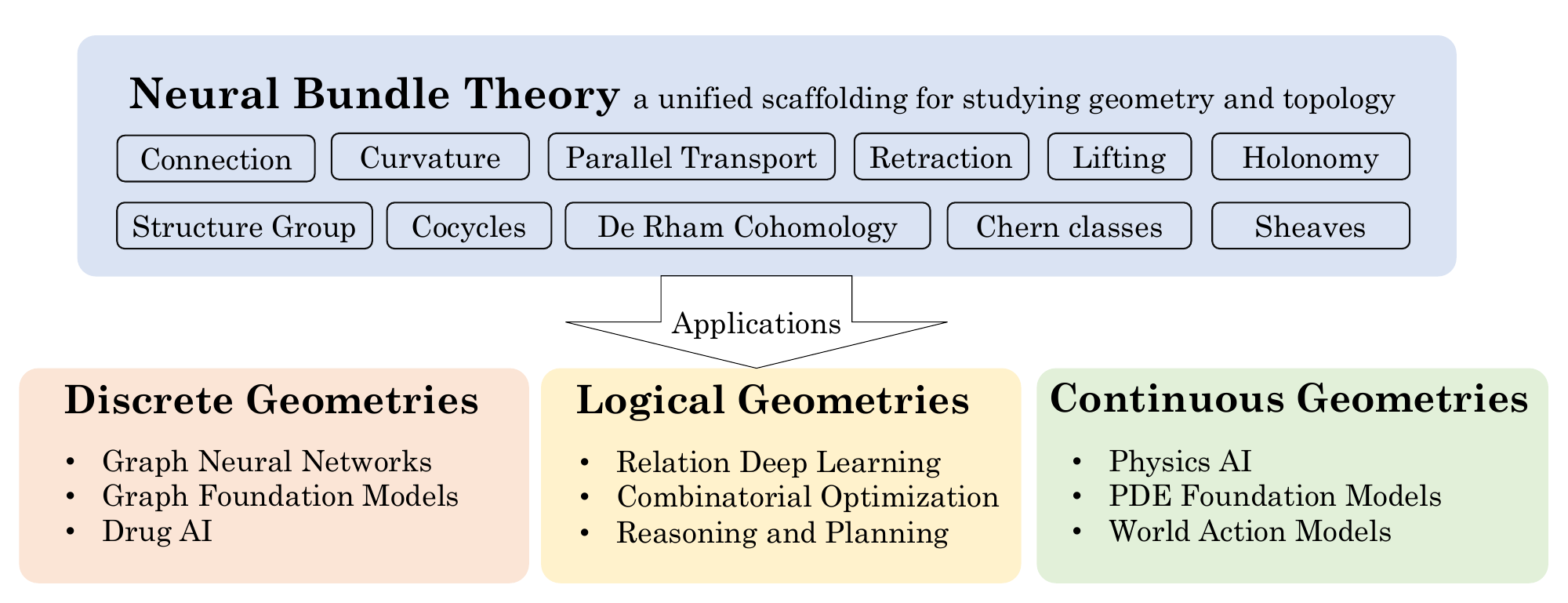}
    \vspace{-0.3in}
    \caption{Theory and Applications of Geometric Morphology Learning.}
    \label{fig1}
    \vspace{-0.2in}
\end{figure}

%\vspace{-0.05in}
\subsubsection*{\textbf{Concepts, Ideas, and Framework.}}
% A core concept is the Geometric Morphology, which refers to  geometric or topological structures, shapes or conformations, such as the geometric shape and boundary of  a fluid field, and the structural regularity of a graph.
% Our idea is concise that the geometric morphology of an object affect its behaviors and, in turn,  learning, reasoning and action should be performed based on such morphology.
% There are two key components in this learning framework, morphology cognition and operator instantiation,
% The morphology cognition is to construct a unified scaffolding  that can cognize any morphology, discrete or continuous, logical or physical.
% Upon such geometric morphology, we are to instantiate learnable operator to study intrinsic geometric structures enable neural networks latent reasoning, deciding and acting based on the geometry.  
% Finally instilling the geometric or topological fabric into  neural networks.
At the core of our framework lies the concept of Geometric Morphology, which encapsulates the intrinsic geometry or topology defining a system, whether manifested as the shape and boundaries of a fluid field or the 3D conformation of a chemical molecule. Our guiding philosophy is simple yet essential: morphology fundamentally dictates the behavior, and correspondingly, learning, reasoning, and action are anchored in this morphological foundation. This framework comprises two key building blocks: Morphology Cognition establishes a unified, universal scaffolding (detailed in the next part) capable of perceiving any morphology across discrete and continuous, logical and physical domains. Building upon this, Operator Instantiation formulates learnable operators to study  intrinsic geometry according to the principle of modern mathematics. Together, these components instill the geometric fabric into neural networks, enabling them to think, reason, decide, and act in accordance with the underlying morphology.

%\vspace{-0.05in}
\subsubsection*{\textbf{Our Neural Bundle Theory.}} 
% We discuss the advancement from Shiing-Shen Chern’s  theory to ours.
% A historical contribution of  Shiing-Shen Chern is the elegant construction of fiber bundles, integrating geometric localities into global topology. 
% Formulated through cohomology theories, 
% Chern classes serve as topological invariant which measures the obstructions to defining global frames on these bundles.
% In spirit of Chern' s theory, we propose Neural Bundles, as the cornerstone of Geometric Morphology Learning, and give the first instantiation [].
% Neural Bundles serve as a unified scaffolding to gain a profound understanding of data geometry. 
% Our  key innovation is to establish fiber bundle with the neural parameterization of local trivialization, where Levi-Civita or Weyl connection exists and Riemannian or pseudo-Riemannian metric is endowed. 
% Our theoretical spectrum covers the neural transportation, curvature, retraction, lifting, and section derived by the bundle construction, illuminating geometric structures at both local and global scales.
% Together with the structure group, Chern classes, cocycles, sheaves, and De Rham cohomology, it forges a principled, unified framework across the geometry and topology through artificial intelligence.
We advance Shiing-Shen Chern’s theory by endowing its learning capabilities. Chern’s seminal work on Fiber Bundle \cite{Chern} integrated local geometric properties with global topology. With fiber bundle theory, Chern classes function as topological invariants, quantifying, through cohomology, the obstructions to defining global frames. In the spirit of Chern’s theory, we introduce Neural Bundles as the cornerstone of Geometric Morphology Learning, and present the first concrete instantiation \cite{NVB}. The core theory and key applications are sketched in Figure \ref{fig1}. Serving as a unified scaffolding, Neural Bundles enable a profound understanding of data geometry. Our key innovation is the construction of fiber bundles via neural parameterization of local trivializations, where Levi-Civita or Weyl connection exists and Riemannian or pseudo-Riemannian metric is endowed. This construction yields a theoretical spectrum encompassing neural transport, curvature, retraction, lifting, and sections, thereby illuminating geometric structures across both local and global scales. By synthesizing structure groups, Chern classes, cocycles, sheaves, and de Rham cohomology, we forge a principled framework that bridges geometry and topology through artificial intelligence.

\subsubsection*{\textbf{Our Pioneering Works.}} 
In the scope of  Geometric Morphology Learning, we model the concept of holonomy with learnable objective to data geometry such as twisting and curvature \cite{sunmulti}, while studying the manifold functional spaces to control structural behaviors at different graph localities \cite{nips25deeper}. 
Another theoretical contribution is that we derive a learnable upper bound of Ollivier-Ricci curvature \cite{icdm23sun}.
%, providing new opportunities for learning on graphs.
We have pioneered neural instantiations of the key concepts in modern geometry, and will create a systematic toolkit and foundations for  Geometric Morphology Learning, allowing for a boarder application of our framework.
Besides, we  pioneer the mixed-curvature modeling for graph foundation models \cite{sun2025riemanngfm}, bridge structural entropy and  hyperbolic geometry \cite{sun2026asil}, and explore optimal transport in Riemannian measure space for generative modeling (Schr\"{o}dinger Bridge) \cite{ijcai25sun}.

%\vspace{-0.05in} 
\subsubsection*{\textbf{Differences to Related Methodologies.}}
%This part briefly elaborates on the differences to related methodologies:
% 2) Manifold learning recovers the low-dimensional manifold  admitting certain geometric priors (e.g., globally connected for Isomap, or the Gaussian for t-SNE) that we are dedicated to break free from.
% 1) Hyperbolic deep learning solely incorporates with a special case of  hyperbolic geometry.
% 2) Manifold learning considers the projection to reduce dimension admitting certain geometric priors, but we emphasize on the morphology of data geometry. 
% 3) Metric learning optimizes  pairwise distance for downstream tasks, while we uncover the task-independent data geometry.
% 4) Geometric deep learning adapts neural networks according to data geometric properties (e.g., equivariance), yet we focus on how to study geometry upon which to reason and act.

%confined to the specific regime of constant negative curvature, 
This part briefly discusses our differences to hyperbolic deep learning, manifold learning, metric learning, and geometric deep learning.
Unlike hyperbolic deep learning, which is confined to hyperbolic spaces, our framework generalizes to arbitrary, variable-curvature geometric structures. In contrast to manifold learning, which focuses on projections for dimensionality reduction, we prioritize the faithful characterization of intrinsic morphology. We also diverge from metric learning, which optimizes pairwise distances merely for specific downstream objectives, by uncovering the task-agnostic, intrinsic geometry of the data. Unlike geometric deep learning, which adapts neural networks to respect data properties (e.g., equivariance), we focus on studying the geometry of the data itself,  laying a geometric  foundation on basis upon which robust reasoning and action can be built.

% \section{Application on Discrete Geometries (Graphs), Logical Geometries (Structural Reasoning) and Continuous Geometries (Physics)}
% \section{Application I: Graphs}
%\section{Application II: Structural Reasoning}
%\section{Application III: Physics AI}

\section{Applications in Discrete, Logical and Continuous Geometries}

Our framework reshapes the landscape of discrete, logical and continuous geometries.
Discrete geometry manifests in graphs, logical geometry in structural reasoning, and continuous geometry in physical systems.

%\vspace{-0.05in}
\subsubsection*{\textbf{Graphs, Laplacian, Spectrum, and Curvatures.}}
Graphs are the discrete approximations of continuous manifolds, making Geometric Morphology Learning fundamental to graph learning and reasoning. Central to graph analysis is  the concept of the graph Laplacian, which mirrors the Laplace-Beltrami operator\footnote{The Laplace-Beltrami operator is a second-order differential operator that explicitly incorporates metric tensor and thus intrinsically governs the heat flow and topological shape of a manifold.}  by quantifying signal smoothness across nodes and governing wave propagation patterns. Leveraging its eigendecomposition, the graph Fourier transform mimics harmonic analysis on  a curved manifold, effectively decomposing graph signals into the spectral (frequency) domain. 
Other important geometric tools include discrete graph curvatures, such as Ollivier–Ricci curvature, which uses optimal transport to rigorously characterize local graph structures.

%\vspace{-0.05in}
\subsubsection*{\textbf{Graph Foundation Model.}}
Identifying generality is vital yet challenging for graph foundation models. Unlike natural languages, which draw on finite vocabularies, graphs exhibit complex, diverse structures that resist principled structural transfer. While existing methods in Euclidean spaces cannot accommodate such complexity, Geometric Morphology Learning paves the way for universal graph understanding. Within this framework, functional spaces over manifolds enable a shift from discrete structures themselves to the more critical behavioral patterns on them. The framework also links the geometric morphology, such as flatness, twisting and curvature, to cross-graph transferability through the topological scaffolding of fiber bundles.

%\vspace{-0.05in}
\subsubsection*{\textbf{Relational Deep Learning}}
Topological structures form the fabric of logical worlds. In relational deep learning, graphs serve as semantic scaffold for multi-table databases, transforming isolated key-value pairs into meaningful entities by defining their relational context; otherwise, the data fragmented. While “JOIN-then-Predict” methods leveraging TablePFNs have emerged \cite{RDL}, Geometric Morphology Learning offers a flexible paradigm aligned with database system principles. Specifically, within the framework of fiber bundles, the covariant derivative\footnote{The covariant derivative corresponds to coboundary operator in sheaf cohomology, understood as relational transportation.}, derived from de Rham cohomology, resolves semantic confusion arising from entity heterogeneity, while the bundle’s curvature assesses the logical consistency across multi-table reasoning paths.
These tools convert abstract topological concepts into learnable operators, thereby optimizing relational reasoning.

%\vspace{-0.05in}
\subsubsection*{\textbf{Combinatorial Optimization, Reasoning, and Planning}}
% Real-world problems are naturally formulated as combinatorial optimization problems such as traveling salesman problems, where state selection is critical, yet standard representation learning or foundation models often fall short. Reinforcement learning relies on step-wise algorithmic execution as feedback, hindering the acquisition of a generalized selection capability. Geometric Morphology Learning provides new opportunities by redefining state transitions through geometric  concepts. Here, retraction acts as a directional mapping onto the manifold, while lifting projects the states backward, elegantly modeling state selection through a Markovian lens, given the existence of a connection. This approach navigates in combinatorial optimization, shedding lights on planning in reality.

Real-world challenges naturally manifest as combinatorial optimization problems, such as the traveling salesman problem, where discrete state selection is paramount, yet representation learning models frequently fall short. Conventional reinforcement learning, which relies on stepwise algorithmic execution as feedback, often struggles to acquire generalized selection capabilities. Geometric Morphology Learning redefines state transitions through rigorous geometric concepts. Specifically, given the existence of a connection, retraction acts as a directional mapping onto the manifold, while lifting projects states backward. This framework geometrically models state selection through a Markovian lens, while navigating the topological space of state trajectories sheds light on robust planning in complex environments.

% Geometry essentially provides a unified perspective for understanding dynamics, energy, momentum, and spacetime, and thus  learning and reasoning with geometry in the physical world via Geometric Morphology Learning opens new opportunities for physics AI and world models.

%\vspace{-0.05in}
\subsubsection*{\textbf{Dynamics, Fields, Lagrangian, and Hamiltonian.}}
% Physical systems can be generally modeled as vector fields. Within the framework of Riemannian manifolds, dynamical systems are elegantly described using geometric structures that naturally incorporate constraints and curved spaces. The configuration space of such a system is represented as a differentiable manifold, where the state of the system evolves along integral curves of specific vector fields. In Lagrangian mechanics, the dynamics are derived from a scalar function, the Lagrangian, typically defined as the difference between kinetic and potential energy on the tangent bundle. The resulting Euler-Lagrange equations dictate the system’s evolution as geodesic flows or their perturbations. Transitioning to Hamiltonian mechanics, the system is formulated on the cotangent bundle, which naturally possesses a symplectic structure. Here, the Hamiltonian, representing the total energy, generates a Hamiltonian vector field via the symplectic form. The resulting Hamiltonian equations describe the flow in phase space, preserving its geometric volume. This geometric perspective not only unifies classical mechanics with differential geometry but also provides a robust foundation for analyzing complex constrained systems, extending seamlessly from classical particle dynamics to modern field theories.
%Physical systems are fundamentally modeled as fields, governed by PDEs on manifolds. It is Riemannian geometry that elevates classical particle dynamics to modern field theories. Geometrically, 

Physics systems are often modeled as PDEs, and it is Riemannian geometry that elevates classical particle dynamics to modern field theories. Within Geometric Morphology Learning,
a field represents a flow over a constrained, curved manifold wherein the system state evolves along the integral curves of specific vector fields. In Lagrangian mechanics, the dynamics are derived from a scalar function, the Lagrangian, typically defined as the difference between kinetic and potential energy on the tangent bundle. The resulting Euler-Lagrange equations dictate the system's evolution as geodesic flows or perturbations thereof. Transitioning to Hamiltonian mechanics, the system is formulated on the cotangent bundle of a symplectic structure. Here, the Hamiltonian, representing the total energy, 
generates a Hamiltonian vector field that preserves the geometric volume of phase space.
Thus, this framework provides a deeper, unified understanding of dynamics, energy and spacetime.
%which naturally possesses a symplectic structure. 
%generates a Hamiltonian vector field via the symplectic form, thereby preserving the geometric volume of the phase space (Liouville's Theorem).

%\vspace{-0.05in}
\subsubsection*{\textbf{World Action Model, Physical Simulation, and Motion Planning}}
% The geometry offers a unified framework for understanding dynamics, energy and spacetime. In this light, Geometric Morphology Learning, learning and reasoning with geometric structures in physical data, essentially opens new avenues for physics AI and world models.
% Geometric constraints construct the reality of physical environments; violating them invites catastrophic consequences. Geometric Morphology Learning empowers environment perception, action modeling, and physical interaction through three progressive layers. First, high-fidelity simulation grounds perception, while scene graphs weave real geometric objects via topology, and fiber bundles unify mass and dynamics beyond 4D spacetime. Second, Christoffel symbols track directional shifts to precisely model actions. Crucially, rather than predicting actions alone, these systems learn representations that couple action generation with future world prediction, seamlessly answering both “what will this action cause” and “how to act to reach a desired future.” Third, topology abstracts these action-environment interactions. Persistent homology captures this structural evolution, while fiber bundles provide a profound angle to study persistent homology from barcode spaces or via their structure groups, forging a robust mathematical bridge between geometry and topology.
Geometric constraints define physical environments, and violating them can invite catastrophic consequences. Geometric Morphology Learning enables environment-action modeling and interaction. Specifically, going beyond the reconstruction of  4D spacetime from visual inputs, fiber bundles further unify kinematic representations to construct high-fidelity environments for studying object topologies, where Christoffel symbols track directional changes to ensure robust action modeling. Crucially, this framework answers  the questions of what this action will cause and how to act to reach a desired future. Coupling  action generation with future world prediction is essentially driven by the persistent homology in topological evolution. In this framework, it can be analyzed using barcode spaces or the structure groups of fiber bundles.

\section{Conclusion}
This paper establishes Geometric Morphology Learning, a novel framework that bridges the distinct realms of Riemannian geometry and deep learning. It addresses a longstanding deficiency in deep learning, namely, its blindness to intrinsic geometry, despite the fundamental role geometry plays in the physical world. 
%Moving beyond reliance on specific geometric priors, we propose to model the intrinsic geometry of objects and to learn with such geometric and topological knowledge. 
This framework transcends predefined priors by learning geometric morphology.
To achieve this, we construct a unified scaffolding of neural bundles through which neural networks  perceive, reason, and act based on intrinsic geometric structures. 
%Equipped with differentiable operators and 
By incorporating advanced geometric concepts, neural networks internalize the governing principles of physics, chemistry, and other abstract structures, aligning machine intelligence with the very fabric of reality. Rather than merely addressing the limitations of deep learning, Geometric Morphology Learning unlocks a deeper, universal language of intelligence, driving transformative progress across diverse scientific and industrial landscapes.

\section*{Acknowledgement}
This work is supported in part by NSFC under grants 62672060.

%%
%% The next two lines define the bibliography style to be used, and
%% the bibliography file.
\bibliographystyle{ACM-Reference-Format}
\bibliography{sample-base}

@inproceedings{ganea2018hyperbolic,
  title={Hyperbolic Neural Networks},
  author={Ganea, Octavian and Becigneul, Gary and Hofmann, Thomas},
  booktitle={Advances in Neural Information Processing Systems},
  volume={31},
  year={2018}
}

@article{mori1988flip,
  title={Flip theorem and the existence of minimal models for 3-folds},
  author={Mori, Shigefumi},
  journal={Journal of the American Mathematical Society},
  volume={1},
  number={1},
  pages={117--253},
  year={1988}
}

@article{yau1974curvature,
  title={Curvature preserving diffeomorphisms},
  author={Yau, Shing-Tung},
  journal={Annals of Mathematics},
  series={2},
  volume={100},
  pages={121--130},
  year={1974}
}

@article{yau1977calabi,
  title={Calabi's conjecture and some new results in algebraic geometry},
  author={Yau, Shing-Tung},
  journal={Proceedings of the National Academy of Sciences},
  volume={74},
  number={5},
  pages={1798--1799},
  year={1977},
}

@article{sun2026asil,
  title     = {ASIL: Augmented Structural Information Learning for Deep Graph Clustering in Hyperbolic Space},
  author    = {Sun, Li and Huang, Zhenhao and Wang, Yujie and Lv, Hongbo and Liu, Chunyang and Peng, Hao and Yu, Philip S.},
  journal   = {IEEE Transactions on Pattern Analysis and Machine Intelligence},
  volume    = {48},
  number    = {6},
  pages     = {6817--6834},
  year      = {2026},
  publisher = {IEEE}
}

@inproceedings{sun2025riemanngfm,
  title={Riemanngfm: Learning a graph foundation model from riemannian geometry},
  author={Sun, Li and Huang, Zhenhao and Zhou, Suyang and Wan, Qiqi and Peng, Hao and Yu, Philip},
  booktitle={Proceedings of the ACM on Web Conference 2025},
  pages={1154--1165},
  year={2025}
}

@inproceedings{sunmulti,
  title={Multi-Domain Riemannian Graph Gluing for Building Graph Foundation Models},
  author={Sun, Li and Huang, Zhenhao and Chen, Silei and Yang, Lanxu and Ye, Junda and Su, Sen and Yu, Philip S},
  booktitle={Proceedings of the 14th ICLR},
  year={2025}
}

@inproceedings{nips25deeper,
  title={Deeper with Riemannian Geometry: Overcoming Oversmoothing and Oversquashing for Graph Foundation Models},
  author={Sun, Li and Huang, Zhenhao and Ming, Zhang and Yu, Philip S},
    booktitle    = {Advances in NeurIPS},
  year={2025}
}

@inproceedings{ijcai25sun,
  author       = {Li Sun and
                  Suyang Zhou and
                  Bowen Fang and
                  Hechuan Zhang and 
                  Junda Ye and
                  Yutong Ye and
                  Philip S. Yu},
  title        = {Trace: Structural Riemannian Bridge Matching for Transferable Source Localization in Information Propagation},
  booktitle    = {Proceedings of the 34th IJCAI},
  pages        = {3308--3316},
  year         = {2025}
}

@inproceedings{icdm23sun,
  title     = {DeepRicci: Self-supervised Graph Structure-Feature Co-Refinement for Alleviating Over-squashing},
  author    = {Li Sun and 
                    Zhenhao Huang and
                    Hua Wu and
                  Junda Ye  and
                   Hao Peng and
                   Zhengtao Yu and
                   Philip S. Yu},
  booktitle    = {Proceedings of the 23rd ICDM},
     pages     = {558--567},
  year      = {2023}
}

@article{GDL,
  author       = {Michael M. Bronstein and
                  Joan Bruna and
                  Taco Cohen and
                  Petar Velickovic},
  title        = {Geometric Deep Learning: Grids, Groups, Graphs, Geodesics, and Gauges},
  year         = {2025}
}

@inproceedings{RDL,
  author       = {Matthias Fey and
                  Weihua Hu and
                  Kexin Huang and
                  Jan Eric Lenssen and
                  Rishabh Ranjan and
                  Joshua Robinson and
                  Rex Ying and
                  Jiaxuan You and
                  Jure Leskovec},
  title        = {Position: Relational Deep Learning - Graph Representation Learning
                  on Relational Databases},
  booktitle    = {Proceedings of 41st ICML},
  volume       = {235},
  pages        = {13592--13607},
  year         = {2024}
}

@inproceedings{NVB,
  author       = {Li Sun and
                  Zhenhao Huang and
                  Yiding Wang and
                  Qian Chen and
                  Pietro Liò and
                  Philip S. Yu},
  title        = {Are Common Substructures Transferable? Riemannian Graph Foundation Model with Neural Vector Bundles},
  booktitle    = {Proceedings of 43rd ICML},
  year         = {2026}
}

@inproceedings{Chern,
  author       = {Shiing-Shen Chern},
  title        = {Differential geometry of fiber bundles},
  booktitle    = {Proceedings of the International Congress of Mathematicians (ICM)},
  year         = {1950},
  volume       = {2},
  pages        = {397--411},
}

@article{RGL,
  author       = {Li Sun and
                  Qiqi Wan and
                  Suyang Zhou and
                  Zhenhao Huang and
                  Philip S. Yu},
  title        = {RiemannGL: Riemannian Geometry Changes Graph Deep Learning},
  year         = {2026},
  journal={CoRR},
}

@BOOK{geo-book,
  author = {Canbin Liang and Bin Zhou},
  title = {Differential Geometry and General Relativity},
  publisher = {Springer},
  year = {2023}
}

@BOOK{manifold-book,
  author = {John M. Lee},
  title = {Introduction to Riemannian Manifolds},
  publisher = {Springer},
  year = {2018}
}

@ARTICLE{Chirality,
  author = {Wu, San and
Chen, Pengquan and
Duan, Meng and
Jiang, Peng-Ying and
Zhou, Qingyang and
Xiang, Shao-Hua and
Houk, K. N. and
Tan, Bin},
  title = {Controlling pyramidal nitrogen chirality by asymmetric organocatalysis},
  journal = {Nature},
  year = {2025},
  volume={647},
  pages={897--905}
}

@ARTICLE{aFold3,
  author = {Abramson, Josh and
Adler, Jonas and
Dunger, Jack and
others},
  title = {Accurate structure prediction of biomolecular interactions with AlphaFold 3},
  journal = {Nature},
  year = {2024},
  volume={630},
  pages={1476--4687}
}

@ARTICLE{material,
  author = {Petra Vojáčková and
  Eric N. Jacobsen},
  title = {An anion-binding approach to enantioselective photoredox catalysis},
  journal = {Science},
    year = {2025},
  volume={390},
  pages={831--836}
}

%%
%% If your work has an appendix, this is the place to put it.
\appendix

\end{document}